\documentclass[sigconf]{acmart}

\setcopyright{none}

\acmDOI{}

\acmISBN{}

\renewcommand\footnotetextcopyrightpermission[1]{}

\usepackage{xspace}

\usepackage{tikz}
\usepackage{pgfplots}
\pgfplotsset{compat=newest}

\begin{document}
\title{AniWho : A Quick and Accurate Way to Classify Anime Character Faces in Images}

\author{Martinus Grady Naftali}
\email{martinus.naftali@binus.ac.id}
\orcid{0000-0003-2695-4913}
\affiliation{%
  \institution{Bina Nusantara University}
  \city{Jakarta}
  \country{Indonesia}
}

\author{Jason Sebastian Sulistyawan}
\email{jason.sulistyawan@binus.ac.id}
\affiliation{%
  \institution{Bina Nusantara University}
  \city{Jakarta}
  \country{Indonesia}
}

\author{Kelvin Julian}
\email{kelvin.julian@binus.ac.id}
\affiliation{%
  \institution{Bina Nusantara University}
  \city{Jakarta}
  \country{Indonesia}
}

\begin{abstract}
\textbf{In order to classify Japanese animation-style character faces, this paper attempts to delve further into the many models currently available, including InceptionV3, InceptionResNetV2, MobileNetV2, and EfficientNet, employing transfer learning. This paper demonstrates that EfficientNet-B7, which achieves a top-1 accuracy of 85.08\%, has the highest accuracy rate. MobileNetV2, which achieves a less accurate result with a top-1 accuracy of 81.92\%, benefits from a significantly faster inference time and fewer required parameters. However, from the experiment, MobileNet-V2 is prone to overfitting; EfficienNet-B0 fixed the overfitting issue but with a cost of a little slower in inference time than MobileNet-V2 but a little more accurate result, top-1 accuracy of 83.46\%. This paper also uses a few-shot learning architecture called Prototypical Networks, which offers an adequate substitute for conventional transfer learning techniques.
}

\end{abstract}


\maketitle
\pagestyle{plain}


\section{Introduction}
\label{sec:introduction}

\noindent
Image recognition is a subbranch of computer vision that aims to identify images based on select features learned by the model \cite{b1}. A large amount of data consisting of images that contain selected objects meant to be recognized is given to the model (e.g., if a model is intended to identify cars, then the data given is many images of vehicles with varieties) \cite{b1, b2}. The model is then trained to define which features make up an object within the image \cite{b3, b4}.

Image recognition is mainly used to perform machine-based tasks such as labeling parts of the image, and performing image searches \cite{b2}. One of the use cases of image recognition is to recognize artists based on their artwork \cite{b5}. Another real-life application of image recognition is mask detection to detect whether the person is appropriately wearing a mask during this pandemic \cite{b6, b7}. One of the topics many people have yet to approach is the face recognition of animated characters such as cartoons especially Japanese animation.

Japanese animation-style cartoons have been snowballing, making billions of dollars in revenue from major streaming platforms such as Netflix and Amazon Prime, resulting in a large audience of at least 35.15 million people in Japan in 2019 \cite{b11}. Besides that, Japanese animation tends to have a wide variety in style and confuses people in identifying the character with only a picture \cite{b12, b13, b14}. We developed this paper based on the large number of people that tend to receive images of unknown Japanese animation-style cartoon images from social media.

Deep learning models are tested to classify Japanese animation-style cartoon character faces. One of the use cases is to develop a recommendation system similar to an application made by Kurt \& Ozkan \cite{b10}, which is an image recommendation system based on the users' input images. The deep learning approach is utilized because a character in anime has many styles of drawing. It is challenging to choose which features are important; The color, shape, and face parts size are sometimes different between different drawing styles. One paper \cite{b15} has tried to do face recognition in anime using manual feature extraction, such as extracting skin color, hair color, and hair quantity. However, it tends to be less accurate at recognizing a similar character or the same character but with a different drawing style.

Other related works \cite{b16} have done experiments with various Convolutional Neural Networks (CNN) that utilize a type of shallownet and a Vision Transformer (VIT) type of architecture. The main objective of this research is to compare several deep learning models' classification accuracies on Japanese animation-style cartoon characters and experiment with different approaches which utilize a few-shot method on prototypical networks. Where our contributions are the following:

\begin{itemize}
\item Top-1, Top-5 Accuracy and inference time evaluation of various transfer learning deep learning models for Face classification of Japanese Animation Cartoon Faces.
\item Empirically proven Prototypical Networks viability for Few-Shot classification of small and imbalanced Japanese Animation Character Faces dataset.
\end{itemize}
This paper is organized as follows: We elaborate on methods and datasets in Section 2. Furthermore, in Section 3, we experiment with transfer learning on various pre-trained convolutional neural network models trained on ImageNet. We freeze all layers except the classification layer to identify which models have the highest accuracy rate. We also utilize a prototypical network to perform few-shot learning, in which the classifier can determine the class of an image, given a small number of examples for each class. Then Section 4 presents the results of the conducted experiments after training the model, whereas Section 5 presents our remarks and future work on this project.

\section{Materials and Method}
\label{sec:Materials and Method}
\noindent In this section, we elaborate more on the various deep learning models used in classifying Japanese cartoon-style cartoon faces. Then, the pre-processing stage for the input images and the parametrization of the machine learning models are compared. Finally,  the dataset we use is explained briefly as well.

\subsection{Inception Network} A deep learning architecture developed by Google is also an extension of the prize-winning inception-v1 \cite{b17}. InceptionV1 introduced the inception module, a module with filters with multiple sizes on the same level \cite{b33}. The inception module removes the hassle of choosing the right kernel size because of the large variation on parts of images. The inception modules are stacked at higher layers after the traditional convolutional layer to improve memory efficiency \cite{b33}. Furthermore, InceptionV1 also adds auxiliary classifiers (discarded at inference time) to solve the problem of deep networks, such as overfitting and vanishing gradient \cite{b33}.

\begin{figure}[!h]
   \centering
   \includegraphics[width=70mm]{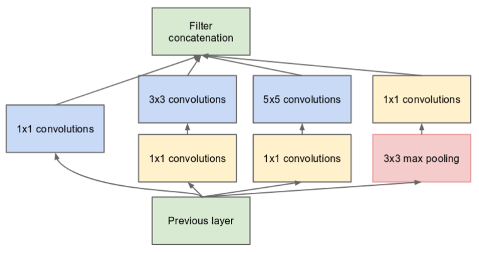}
    \caption{Inception Module. Figure taken from~\cite{b33}}
    \label{fig1}
\end{figure}

\begin{figure}[!h]
   \centering
   \includegraphics[width=\linewidth]{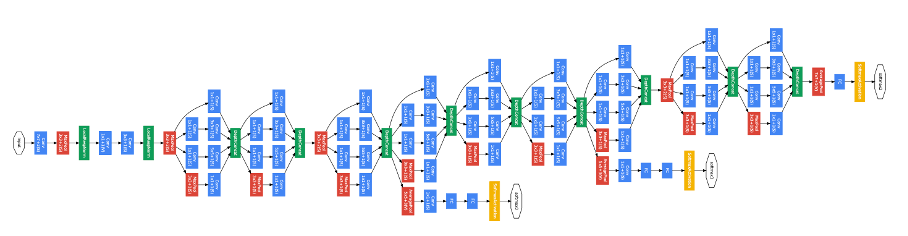}
    \caption{InceptionV1 Archtecture. Figure taken from~\cite{b33}}
    \label{fig2}
\end{figure}

The next version, inceptionV2, reduces the computational complexity using spatial factorization \cite{b18}. Spatial factorization factorizes filter size convolution into asymmetric convolutions \cite{b18}. Additionally, filter banks in the inception module were expanded to prevent excessive reduction in dimension \cite{b18}. 

\begin{figure}[!h]
   \centering
   \includegraphics[width=50mm]{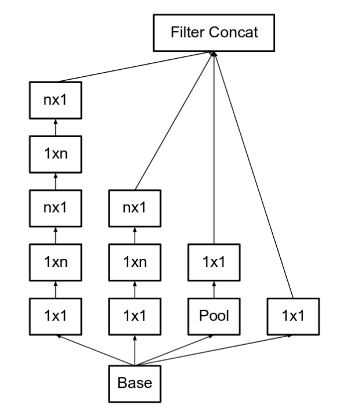}
    \caption{Inception module after the spatial factorization. Figure taken from~\cite{b18}}
    \label{fig3}
\end{figure}

\begin{figure}[!h]
   \centering
   \includegraphics[width=50mm]{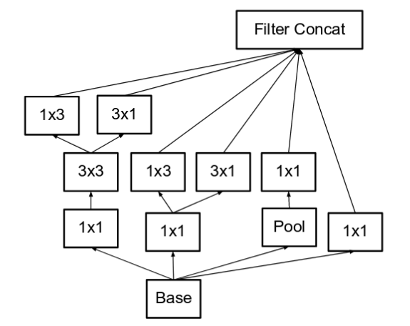}
    \caption{Inception module with expanded filter banks. Figure taken from~\cite{b18}}
    \label{fig4}
\end{figure}

Finally, inceptionV3 further optimizes the accuracy and performance outcome by factorizing a 7x7 convolution, using RMSProp Optimizer to speed up gradient descent, and regularizing a component called label smoothing to prevent overfitting \cite{b18}. Previously there were two auxiliary classifiers in the network. However, because it is analyzed that it only contributes little until near the end of the training process, it was reduced to one \cite{b18}. In inceptionV3, the auxiliary classifier has a different purpose, which is to be used as a regularizer \cite{b18}. In other works, the InceptionV3 model is mainly used for transfer learning, and then the model is also trained once more with the specific datasets used in the experiment \cite{b20,b21,b22}. 

\begin{figure}[!h]
   \centering
   \includegraphics[width=\linewidth]{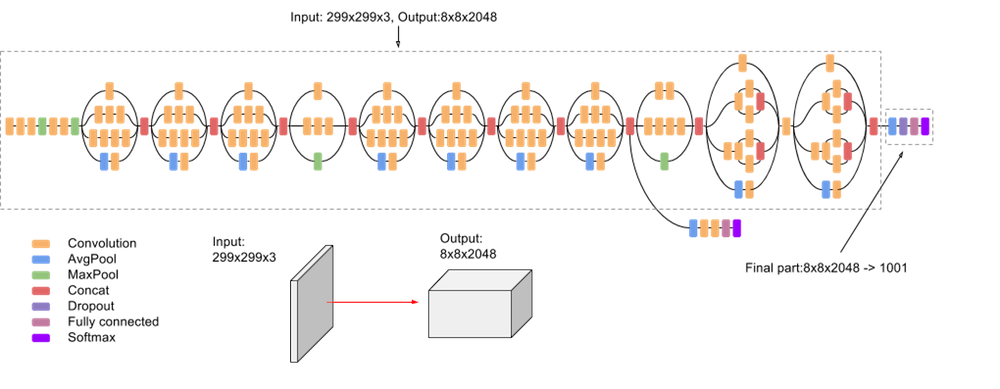}
    \caption{High-level Diagram of the Inceptionv3 Architecture. Figure taken from~\cite{b19}}
    \label{fig5}
\end{figure}

\subsection{InceptionResNetV2} InceptionResnetV2 modified the inception network to avoid the vanishing gradient problem and accelerate training. The residual connection was added to introduce a skip connection which connects the output of the previous convolutional layer in the inception module to the several next ones \cite{b24}. InceptionResnetV2 has an inception and reduction module; the inception module is divided into three different variants of blocks (inception a, b, and c) and two variations of reduction module (reduction a and b) \cite{b24}. Reduction modules are used to change the width and height of the grid from the inception module \cite{b24}. 
Other works tend to utilize the model for goals of object recognition within images \cite{b23} through its feature extraction capabilities. The following Figure 6 illustrates the schematic diagram of the InceptionResNetV2 architecture.

\begin{figure}[!h]
	\begin{center}
  		\includegraphics[width=60mm]{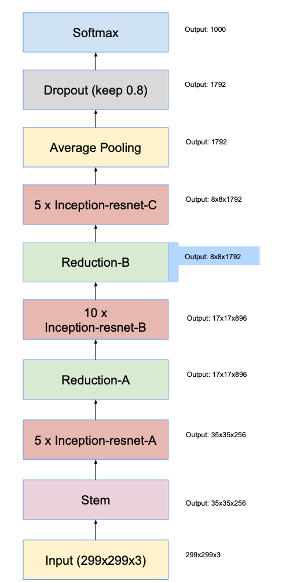}
  	\end{center}
  	\caption{Schema Diagram of the InceptionResNetV2 network. Figure taken from~\cite{b24}}
 	 \label{fig6}
\end{figure}

\begin{figure}[!h]
	\begin{center}
  		\includegraphics[width=50mm]{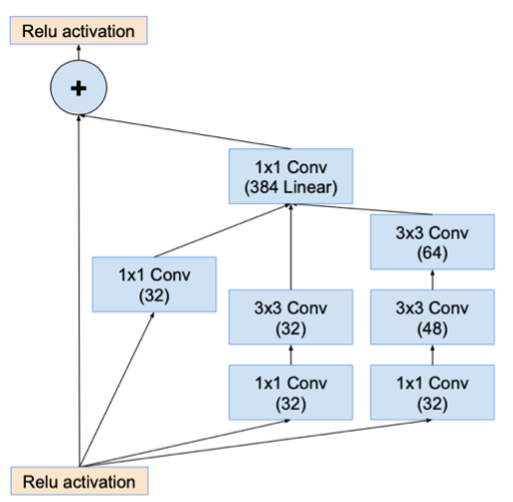}
  	\end{center}
  	\caption{Schema Diagram of Inception Module-A of the InceptionResNetV2 network. Figure taken from~\cite{b24}}
 	 \label{fig7}
\end{figure}

\begin{figure}[!h]
	\begin{center}
  		\includegraphics[width=50mm]{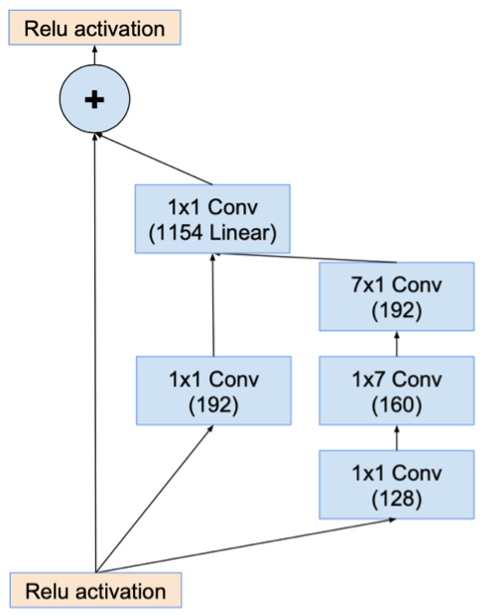}
  	\end{center}
  	\caption{Schema Diagram of Inception Module-B of the InceptionResNetV2 network. Figure taken from~\cite{b24}}
 	 \label{fig8}
\end{figure}

\begin{figure}[!h]
	\begin{center}
  		\includegraphics[width=50mm]{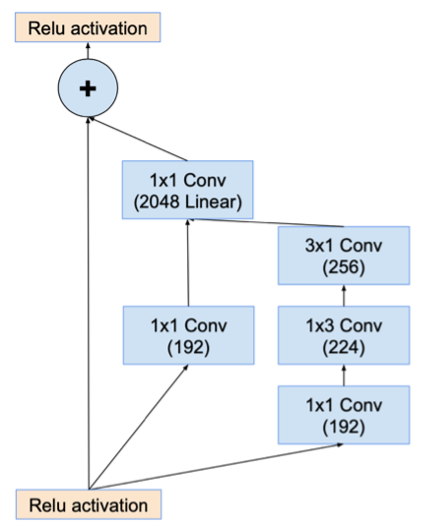}
  	\end{center}
  	\caption{Schema Diagram of Inception Module-C of the InceptionResNetV2 network. Figure taken from~\cite{b24}}
 	 \label{fig9}
\end{figure}

\begin{figure}[!h]
	\begin{center}
  		\includegraphics[width=50mm]{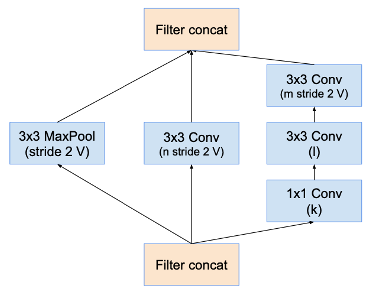}
  	\end{center}
  	\caption{Schema Diagram of Reduction Module-A of the InceptionResNetV2 network. Figure taken from~\cite{b24}}
 	 \label{fig10}
\end{figure}

\begin{figure}[!h]
	\begin{center}
  		\includegraphics[width=55mm]{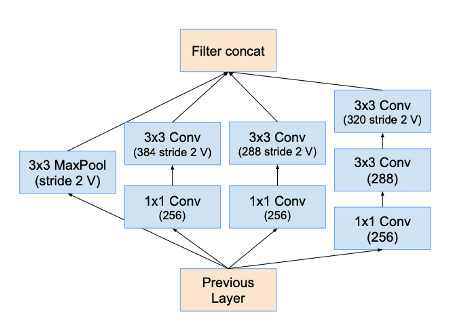}
  	\end{center}
  	\caption{Schema Diagram of Reduction Module-B of the InceptionResNetV2 network. Figure taken from~\cite{b24}}
 	 \label{fig11}
\end{figure}

\subsection{MobileNet} 
MobileNetV1 uses depthwise separable convolution, which reduces the computational complexity significantly compared to standard convolution but slightly reduces accuracy \cite{b35}. Depthwise separable convolution is divided into two parts which are depthwise and pointwise convolution \cite{b35}. Depthwise convolution convolves each input channel in a different single filter, and the pointwise layer combines the output of depthwise convolution with 1x1 convolution \cite{b35}. Batchnorm and ReLu activation functions are added after each depthwise and pointwise layer 

\begin{figure}[!h]
	\begin{center}
  		\includegraphics[width=45mm]{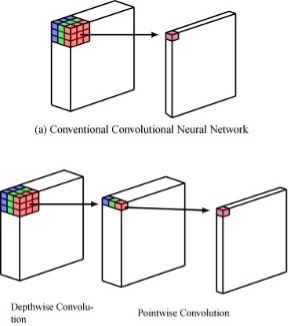}
  	\end{center}
  	\caption{Conventional Convolutional Neural Network vs Depthwise Separable Convolutional Neural Network. Figure taken from~\cite{b36}}
 	 \label{fig12}
\end{figure}

\begin{figure}[!h]
	\begin{center}
  		\includegraphics[width=30mm]{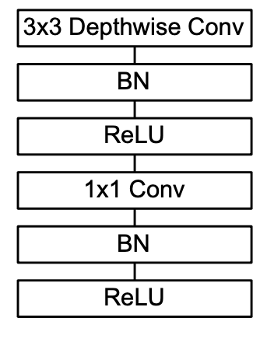}
  	\end{center}
  	\caption{Depthwise and Pointwise layers followed by batchnorm and ReLU. Figure taken from~\cite{b35}}
 	 \label{fig13}
\end{figure}

The second version, MobilenetV2, follows the inverted version of the residual block structure called inverted residual \cite{b25}. Inverted residual, like the original residual, also has a skip connection (connects between bottleneck layers) to avoid vanishing gradient but with considerably more memory efficiency \cite{b25, b26, b27}. The interpretation of the inverted residual in MobilenetV2 is called Bottleneck Residual Block. This block is divided into three parts: expansion, depthwise, and projection layer \cite{b25}. The expansion layer is a 1x1 convolution that expands the number of channels; the purpose is that so filtering in the depthwise layer is done on a much higher dimension \cite{b25}. 1 x 1 Linear Bottleneck layer is then used to make the channel smaller again, so computational complexity between blocks is cheaper and prevents non-linearities from destroying too much information \cite{b25}. The final architecture is described in figure 16. 

\begin{figure}[!h]
	\begin{center}
  		\includegraphics[width=70mm]{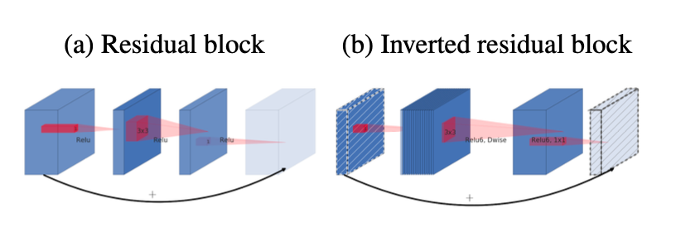}
  	\end{center}
  	\caption{Residual Block and Inverted Residual Block. Figure taken from~\cite{b25}}
 	 \label{fig14}
\end{figure}

\begin{figure}[!h]
	\begin{center}
  		\includegraphics[width=50mm]{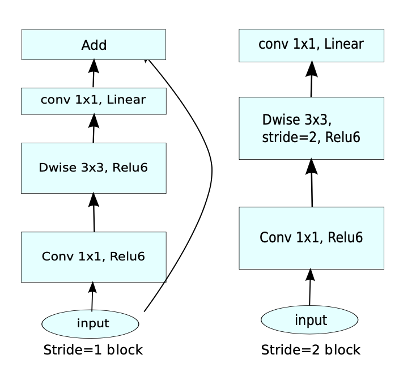}
  	\end{center}
  	\caption{Bottleneck Residual Block. Figure taken from~\cite{b25}}
 	 \label{fig15}
\end{figure}

\begin{figure}[!h]
	\begin{center}
  		\includegraphics[width=70mm]{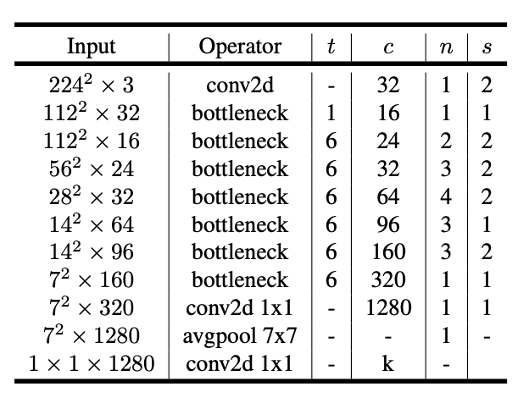}
  	\end{center}
  	\caption{MobilenetV2 Architecture. Figure taken from~\cite{b25}}
 	 \label{fig16}
\end{figure}

\subsection{EfficientNet} The EfficientNet is a family of optimized models for FLOPs and parameters efficiency \cite{b28}. EfficientNet implemented a compound scaling method that balances different types of model scaling in the network using compound coefficient $\phi$ to uniformly scale network width, depth, and resolution \cite{b28}. Furthermore, efficientnet is also constructed using an inverted residual block introduced in mobilenetv2, called MBConv \cite{b25}. MBConv is added using squeeze and excite optimization, which improves channel independencies by adding content aware mechanism in each channel in the convolutional block \cite{b37}. Some works \cite{b28,b29} have obtained that the EfficientNetB7 version is the most accurate. The baseline architecture of efficientnet is described in figure 18.

\begin{figure}[!h]
  \includegraphics[width=50mm]{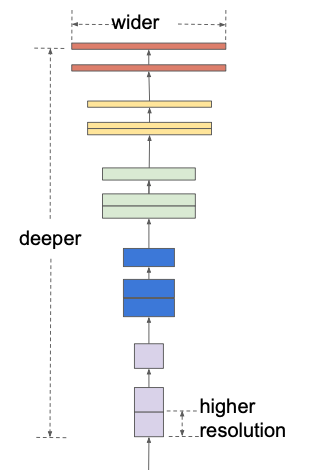}
  \caption{Compound Scaling.  Figure taken from~\cite{b28}}
  \label{fig17}
\end{figure}

\begin{figure}[!h]
  \includegraphics[width=\linewidth]{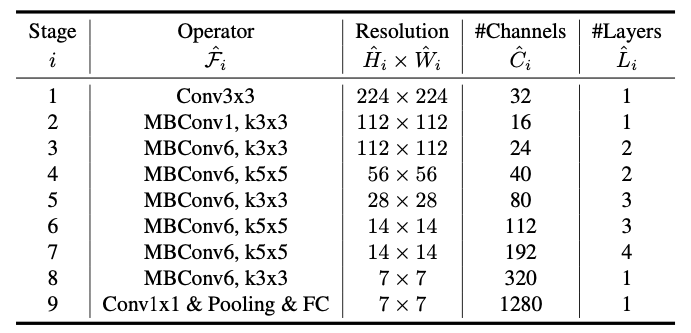}
  \caption{baseline EfficientNet Architecture (EfficientNet-B0).  Figure taken from~\cite{b28}}
  \label{fig18}
\end{figure}

\subsection{Prototypical Network} The Prototypical Network is a few-shot learning framework that performs metric learning to perform classification using the computed distances to prototype representations of each class \cite{b31}. The Prototypical Networks use a CNN with a chopped classification head as a backbone to project support and query images into feature space \cite{b31}. 

The mean of the embeddings of images from a class in support-set forms the Class prototypes \cite{b31}. The softmax of the Euclidian distance between each query image's embeddings and the class prototype in support-set embeddings is used to classify the images \cite{b31}. The CNN is trained using cross-entropy loss to update model's parameters \cite{b31}.

Based on other work \cite{b32}, Prototypical Network has several drawbacks leading to undesired classification performance. A Prototypical Network only considers the mean vectors, creating misclassification when the variance in the support-set is relatively large. The following Figure 19 below gives an illustration of a Prototypical Network.

\begin{figure}[!h]
	\begin{center}
  		\includegraphics[width=50mm]{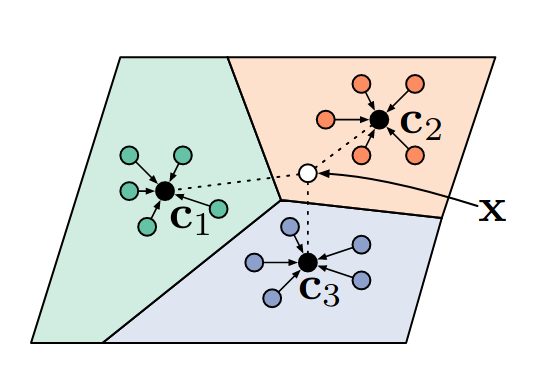}
  	\end{center}
  	\caption{Prototypical Network. Ci are the computed mean of the embedded support-set of each class. X is the query point that needs to be classified. Figure taken from~\cite{b32}}
  	\label{fig19}
\end{figure}

\subsection{Image Pre-Processing and Data Augmentation} The process of image pre-processing consists of many operations, such as rescaling and cropping. The process of image pre-processing itself is done to enable the input images to be used with the corresponding input layers based on the current model used. 

\subsection{Parametrization of Image Classification Algorithms and Training Details} In image classification, most models undergo a training process where the model is given a collection of input images based on their corresponding ground truth boxes for each of their classes. In our approach to building our model, we utilize the transfer learning method from the previously selected models: the InceptionV3, InceptionResNetV2, MobileNetV2, EfficientNetB0, EfficientNetB7, and a few-shot method, with a Prototypical Network framework.

In this paper, to compare the transfer learning models used based on the Japanese Styled Characters Images, the main metrics that are compared consist of; train top-1 accuracy, train top-5 accuracy, validation top-1 accuracy, validation top-5 accuracy, test top-1 accuracy, test top-5 accuracy, and inference time. Meanwhile, for prototypical networks, only accuracy is used for comparison.

\subsection{Utilized Dataset} The dataset used in this paper consists of 9738 images and 130 character classes, with approximately 75 images in each class extracted from the Danbooru website, a board developed by the Japanese animation-style cartoon community for image hosting and collaborative tagging \cite{b15}.


\section{Experiments}
\label{sec:experiments}
\begin{figure}[!h]
  \includegraphics[width=\linewidth]{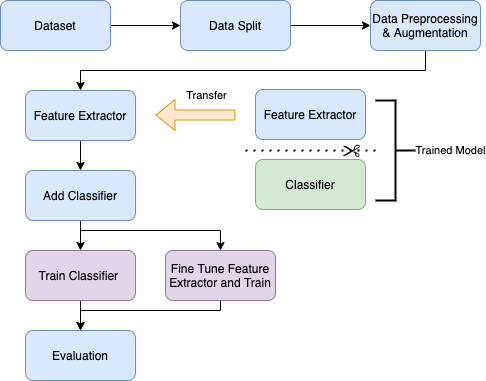}
  \caption{Transfer Learning Experiment Flow.}
  \label{fig20}
\end{figure}

\subsection{Transfer Learning Experiment}
In this experiment, the dataset images are cropped so that only the characters’ face is visible and rescaled into a size of 96 × 96 pixels. The dataset is split into 90\% training and 10\% test sets. In this whole experiment, the TensorFlow framework was used. Furthermore, a TensorFlow preprocessing module was used to split the training set into 18\% validation and 72\% training set. Then, the images are preprocessed and normalized to the scale of 1 / 255. Additionally, the training dataset was augmented during training with the following settings:

\begin{table}[!h]
\caption{Dataset Augmentation Settings}
\label{table}
\setlength{\tabcolsep}{3pt}
\begin{tabular}{|p{100pt}|p{115pt}|}
\hline
Augmentation& 
Value 
\\
\hline
RandomRotation& 
0.4 \\
RandomTranslation& 
0, 0.2 \\
RandomTranslation& 
0.2, 0 \\
RandomZoom& 
0.2, 0.2 \\
RandomFlip& 
mode="horizontal\_and\_vertical" \\
RandomContrast& 
0.2 \\
\hline
\end{tabular}
\label{tab1}
\end{table}

To compare machine learning models for facial classification of Japanese animation-style cartoon images with transfer learning, models from InceptionV3, InceptionResNetV2, MobileNetV2, EfficientNetB0, and EfficientNetB7 with pre-loaded weights trained on the ImageNet dataset were instantiated. All classification layers of the models were not included for feature extraction.

A classification head was added on top of the network, which included a couple of new layers with the hub.KerasLayer, GlobalAveragePooling2D layer was already included; this layer converts features to a single 1280-element vector per image. A Dropout layer was also added to prevent overfitting during training by randomly setting input units in the network to 0. Finally, a Dense layer with an activation function inside was added to convert features into a single prediction per image.

The base learning rate for training is 0.005, and the number of epochs is 100. Stochastic gradient descent (SGD) optimizer was used with a momentum of 0.9 to optimize training;  the decay rate is calculated with the learning rate divided by epoch. Furthermore, models were compiled with the Categorical\_Crossentropy loss function and metrics settings, including accuracy and topKCategoricalAccuracy with a k value of 5. 

EarlyStopping, which monitored validation loss and patience value of 6, was applied to reduce training time. Furthermore, ReduceLROnPlateau was also applied to monitor the validation loss with a factor value of 0.2, patience value of 3, and minimum learning rate of  0.00001. This learning rate reduction is also crucial when the current learning rate stops improving during training. 

For evaluation, each model used was tested with a batch size of 16 and 32, dropout layer in the classification head with rate values of 0.2, 0.5, and 0.8. Finally, the models were also tested with fine-tuning and without fine-tuning. Top-1, Top-5 Accuracy, and inference time of each model were evaluated on a Personal Computer(PC) with 8th Gen. Intel® Core™ i5-8600K Processor 3.60GHz CPU, NVIDIA® GeForce® GTX 1050 TI GPU, and 16GB DDR4 3200MHz RAM.

\begin{figure}[!h]
	\begin{center}
  		\includegraphics[width=72mm]{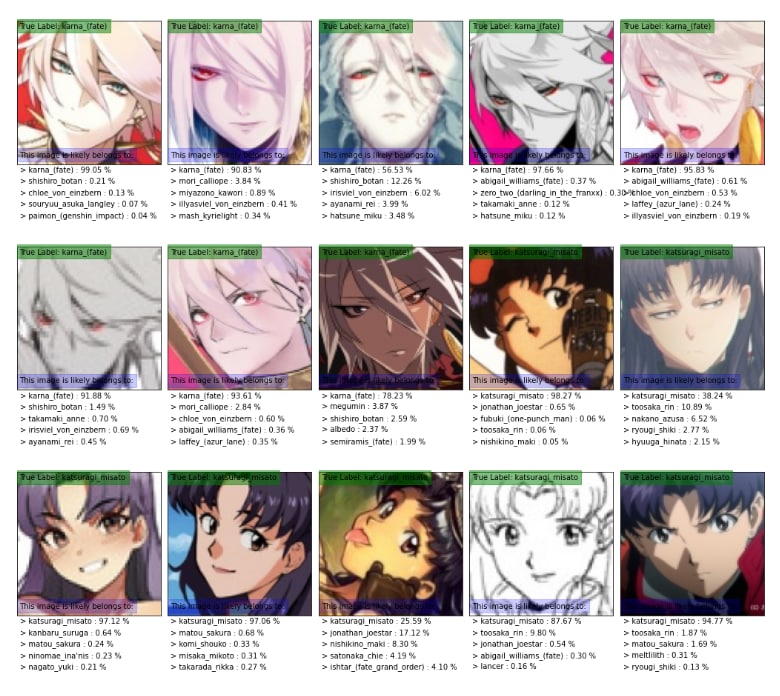}
  	\end{center}
  	\caption{Top-5 Accuracy means it returns true if one of top 5 predicted labels is the correct label.}
  	\label{fig21}
\end{figure}

\subsection{Few-Shot Prototypical Network Experiment}

\begin{figure}[!h]
  \includegraphics[width=\linewidth]{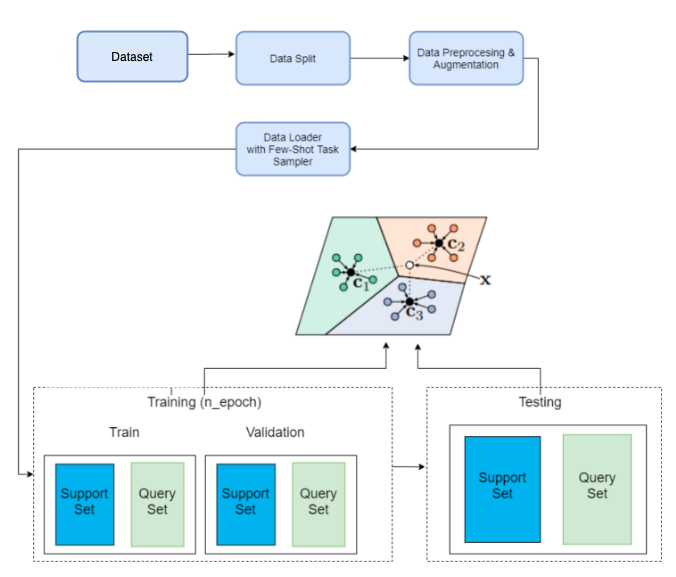}
  \caption{Few-Shot Prototypical Network Experiment Experiment Flow.}
  \label{fig22}
\end{figure}

In the Few-Shot classification experiment, the dataset preparation was similar to the transfer learning experiment. However, the split process was different; the 130 classes in the dataset were divided into 90 classes for the training set, 20 classes for the validation set, and 20 classes for the test set where none of the sets have the same class in them. Pytorch with an easy-few-shot-learning library was used for this few-shot classification with a prototypical network experiment.

Each set was then preprocessed and augmented using the easyset module. For the training set, images were randomly resized and cropped depending on the image size; then, color jitter was applied with a brightness of 0.4, a contrast of 0.4, and saturation of 0.4. Random horizontal flip was also applied, and finally, the images were converted to tensors and normalized. For validation and test set, different settings were applied to the images. First, the images were cropped and centered, and finally, the images were converted to tensors and normalized.

\begin{figure}[!h]
	\begin{center}
  		\includegraphics[width=70mm]{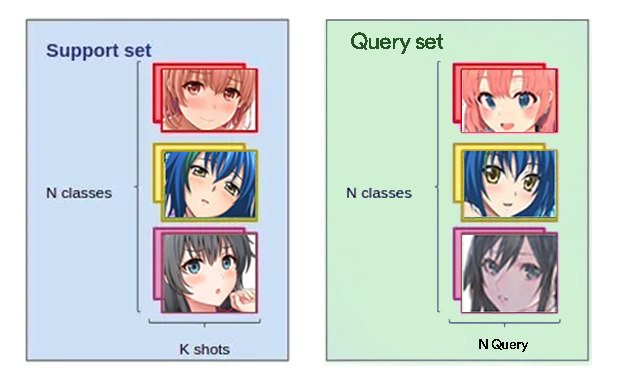}
  	\end{center}
  	\caption{Left is support set and right is query set.}
  	\label{fig23}
\end{figure}

Each set was loaded to the DataLoader module using a customized TaskSampler provided by the easy-few-shot-learning library to feed a few-shot classification task to the model. Each set then will be divided into support sets and query sets with a combination of several classes, K shot or several images per class in the support set, several tasks, and N query or number of images per class in the query set. All the experiments were tested with ten images of the query and 100 tasks per epoch training, 100 tasks validation, and 1000 tasks testing. 4 Experiments were conducted with 5-class 5-shot, 5-class 1-shot, 20-class 5-shot, and 20-class 1-shot.

\begin{figure}[!h]
	\begin{center}
  		\includegraphics[width=50mm]{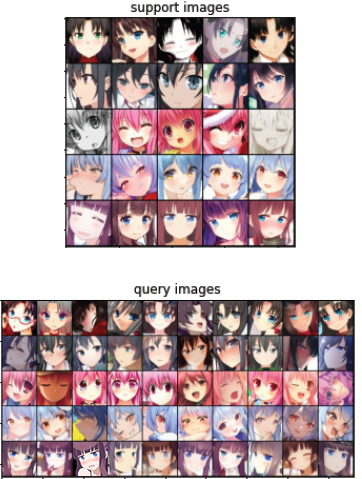}
  	\end{center}
  	\caption{5-class 5-shot on support set and 10 query on query set.}
  	\label{fig24}
\end{figure}

The training process in a few-shot is called Meta-Training. In Meta-Training, Prototypical Networks need the backbone to project support and query images into feature space. This experiment uses custom ResNet architecture with pre-trained weights provided by the easy-few-shot-learning library for the backbone. 100 epochs, CrossEntropyLoss, and a base learning rate of 0.01 were set for training. To optimize the training, SGD was used with a momentum of 0.9 and a weight decay of 0.0005. A learning rate scheduler was also applied using MultiStepLR with a gamma of 0.1 and epoch milestones of 50 and 80. 

\begin{figure}[!h]
	\begin{center}
  		\includegraphics[width=70mm]{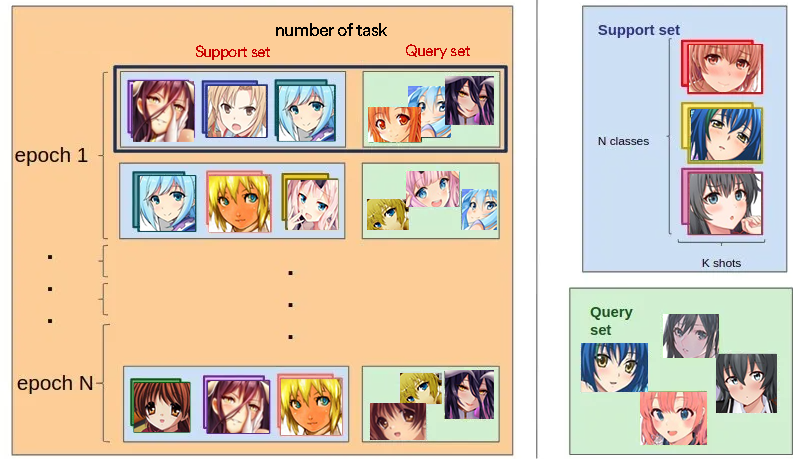}
  	\end{center}
  	\caption{Meta-Training.}
  	\label{fig25}
\end{figure}

\section{Results and Discussion}
\label{sec:ResultsandDiscussion}
\subsection{Transfer Learning}
The following tables below consist of the results based on each machine learning model utilized with a description of the hyperparameter values. 

\begin{table}[!h]
\caption{InceptionV3 Classification Results}
\label{table}
\setlength{\tabcolsep}{3pt}
\renewcommand{\arraystretch}{1.1}
\resizebox{.48\textwidth}{!}{
\begin{tabular}{|c|c|c|c|c|c|c|c|}
\hline
Model & Batch Size & Fine Tune & Dropout Rate & Accuracy
(\%) & Train & Valid & Test \\
\hline
{ InceptionV3 } & { 32 } & { FALSE } & { 0.2 } & top-1 & 22.84 & 25.52 & 26.37 \\
& & & & top-5 & 54.36 & 54.36 & 54.48 \\
\cline{4-8}
& & & { 0.5 } & top-1 & 64.29 & 35.68 & 38.94 \\
& & & & top-5 & 90.2 & 66.02 & 65.77 \\
\cline{4-8}
& & & { 0.8 } & top-1 & 36.26 & 35.81 & 36.57 \\
& & & & top-5 & 67.24 & 64.13 & 65.06 \\
\cline{3-8}
& & { TRUE } & { 0.2 } & top-1 & 100 & 75.98 & \textbf{76.73} \\
& & & & top-5 & 100 & 91.15 & \textbf{90.87} \\
\cline{4-8}
& & & { 0.5 } & top-1 & 8.98 & 12.96 & 12.8 \\
& & & & top-5 & 32.8 & 39.19 & 40.9 \\
\cline{4-8}
& & & { 0.8 } & top-1 & 2.35 & 2.6 & 4.04 \\
& & & & top-5 & 10.94 & 12.5 & 15.7 \\
\cline{2-8}
& { 16 } & { FALSE } & { 0.2 } & top-1 & 23.52 & 27.19 & 28.01 \\
& & & & top-5 & 50.7 & 56.38 & 57.75 \\
\cline{4-8}
& & & { 0.5 } & top-1 & 68.17 & 34.92 & 35.9 \\
& & & & top-5 & 92.24 & 64.56 & 63.91 \\
\cline{4-8}
& & & { 0.8 } & top-1 & 37.05 & 33.51 & 34.94 \\
& & & & top-5 & 68.19 & 64.56 & 63.33 \\
\cline{3-8}
& & { TRUE } & { 0.2 } & top-1 & 59.69 & 50.77 & 50.63 \\
& & & & top-5 & 87.94 & 74.1 & 77.57 \\
\cline{4-8}
& & & { 0.5 } & top-1 & 37.15 & 39.56 & 37.82 \\
& & & & top-5 & 71.32 & 68.11 & 71.32 \\
\cline{4-8}
& & & { 0.8 } & top-1 & 1.83 & 1.68 & 2.69 \\
& & & & top-5 & 8.89 & 8.76 & 11.65 \\
\hline
\end{tabular}}
\label{tab2}
\end{table}

The result above on InceptionV3 shows that overall fine-tuning gives better test results with a dropout rate of 0.2. This is because fine-tuning unfreezes a few top layers from the frozen base model of the network, enabling training on those layers. While a dropout rate of 0.5 and 0.8 does not always gives a better result when fine-tuned because this model might be prone to lose more information. Overall, batch size only made little difference in test result value between batch sizes 32 and 16 except on batch size 32 with fine-tuning enabled and a dropout rate of 0.2. This is because the higher the batch size, the lesser the noise in the gradients, which makes the gradient estimate better. However, this caused overfitting since the training resulted in a value of 100\%. This is because more layers were exposed to training while only a dropout of 0.2 was used.

\begin{table}[!h]
\caption{InceptionResNetV2 Classification Results}
\label{table}
\setlength{\tabcolsep}{3pt}
\renewcommand{\arraystretch}{1.1}
\resizebox{.48\textwidth}{!}{
\begin{tabular}{|c|c|c|c|c|c|c|c|}
\hline
Model & Batch Size & Fine Tune & Dropout Rate & Accuracy(\%) & Train & Valid & Test \\
\hline
{ InceptionResNetV2 } & { 32 } & { FALSE } & { 0.2 } & top-1 & 88.44 & 44.01 & 45.91 \\
& & & & top-5 & 98.56 & 71.35 & 71.32 \\
\cline{4-8}
& & & { 0.5 } & top-1 & 61.12 & 44.14 & 45.91 \\
& & & & top-5 & 86.96 & 69.53 & 71.13 \\
\cline{4-8}
& & & { 0.8 } & top-1 & 37.19 & 40.49 & 42.35 \\
& & & & top-5 & 67.92 & 67.12 & 68.82 \\
\cline{3-8}
& & { TRUE } & { 0.2 } & top-1 & 100 & 73.44 & 70.48 \\
& & & & top-5 & 100 & 90.23 & 89.89 \\
\cline{4-8}
& & & { 0.5 } & top-1 & 70.45 & 75.39 & \textbf{74.98} \\
& & & & top-5 & 89.75 & 90.89 & \textbf{90.76} \\
\cline{4-8}
& & & { 0.8 } & top-1 & 70.45 & 69.08 & 67.66 \\
& & & & top-5 & 89.75 & 89.19 & 88.93 \\
\cline{2-8}
& { 16 } & { FALSE } & { 0.2 } & top-1 & 28.28 & 32.22 & 31.76 \\
& & & & top-5 & 56.16 & 62.18 & 62.85 \\
\cline{4-8}
& & & { 0.5 } & top-1 & 18.6 & 28.03 & 29.84 \\
& & & & top-5 & 43.21 & 57.47 & 59.00 \\
\cline{4-8}
& & & { 0.8 } & top-1 & 10.31 & 23.78 & 24.06 \\
& & & & top-5 & 27.95 & 50.97 & 50.53 \\
\cline{3-8}
& & { TRUE } & { 0.2 } & top-1 & 73.7 & 65.66 & 65.83 \\
& & & & top-5 & 94.6 & 88.02 & 88.93 \\
\cline{4-8}
& & & { 0.5 } & top-1 & 27.99 & 33.7 & 33.3 \\
& & & & top-5 & 58.54 & 64.95 & 65.06 \\
\cline{4-8}
& & & { 0.8 } & top-1 & 3.6 & 5.73 & 7.12 \\
& & & & top-5 & 15.01 & 24.48 & 26.28 \\
\hline
\end{tabular}}
\label{tab3}
\end{table}

InceptionResNetV2 is better when fine-tuned while using dropout rates of 0.2 and 0.5. This is because the 0.8 rate causes more nodes to be dropped out. When fine-tuned, this model also shows that unfreezing a few top layers of the model enables more layers to be trained for this specific task. Furthermore, the number of batches also shows a significant difference because of the same reason that happened in inceptionV3.

\begin{table}[!h]
\caption{MobileNetV2 Classification Results}
\label{table}
\setlength{\tabcolsep}{3pt}
\renewcommand{\arraystretch}{1.1}
\resizebox{.48\textwidth}{!}{
\begin{tabular}{|c|c|c|c|c|c|c|c|}
\hline
Model & Batch Size & Fine Tune & Dropout Rate & Accuracy(\%) & Train & Valid & Test \\
\hline
{ MobileNetV2 } & { 32 } & { FALSE } & { 0.2 } & top-1 & 95.09 & 58.53 & 57.88 \\
& & & & top-5 & 99.75 & 81.58 & 82.69 \\
\cline{4-8}
& & & { 0.5 } & top-1 & 82.52 & 60.42 & 59.33 \\
& & & & top-5 & 97.68 & 83.4 & 83.37 \\
\cline{4-8}
& & & { 0.8 } & top-1 & 48.3 & 59.24 & 59.42 \\
& & & & top-5 & 77.23 & 81.97 & 81.92 \\
\cline{3-8}
& & { TRUE } & { 0.2 } & top-1 & 97.77 & 81.12 & 80.19 \\
& & & & top-5 & 100 & 93.16 & 92.69 \\
\cline{4-8}
& & & { 0.5 } & top-1 & 100 & 82.03 & \textbf{81.92} \\
& & & & top-5 & 100 & 93.88 & \textbf{94.62} \\
\cline{4-8}
& & & { 0.8 } & top-1 & 97.44 & 76.69 & 75.58 \\
& & & & top-5 & 99.86 & 91.73 & 90.67 \\
\cline{2-8}
& { 16 } & { FALSE } & { 0.2 } & top-1 & 97.08 & 59.41 & 57.79 \\
& & & & top-5 & 99.93 & 81.77 & 81.63 \\
\cline{4-8}
& & & { 0.5 } & top-1 & 79.13 & 59.73 & 60.38 \\
& & & & top-5 & 96.52 & 82.6 & 82.4 \\
\cline{4-8}
& & & { 0.8 } & top-1 & 49.19 & 59.34 & 59.33 \\
& & & & top-5 & 79.37 & 82.28 & 82.12 \\
\cline{3-8}
& & { TRUE } & { 0.2 } & top-1 & 100 & 82.99 & 81.54 \\
& & & & top-5 & 100 & 93.94 & 93.17 \\
\cline{4-8}
& & & { 0.5 } & top-1 & 99 & 83.31 & 82.12 \\
& & & & top-5 & 100 & 94.33 & 93.85 \\
\cline{4-8}
& & & { 0.8 } & top-1 & 94.69 & 77.77 & 77.4 \\
& & & & top-5 & 99.52 & 92.59 & 93.17 \\
\hline
\end{tabular}}
\label{tab4}
\end{table}

Overall, with MobileNetV2, batch size does not make a significant difference in test results; furthermore, fine-tuning results were better because of the same reason as the previous model. This model is also suitable for using a dropout rate of 0.2 and 0.5, the same as InceptionResNetV2, because, with a rate of 0.8, there might be too many dropped nodes. It is also seen that this model is prone to overfitting because most training results are near 100\%. 

\begin{table}[!h]
\caption{EfficientNetB0 Classification Results}
\label{table}
\setlength{\tabcolsep}{3pt}
\renewcommand{\arraystretch}{1.1}
\resizebox{.48\textwidth}{!}{
\begin{tabular}{|c|c|c|c|c|c|c|c|}
\hline
Model & Batch Size & Fine Tune & Dropout Rate & Accuracy(\%) & Train & Valid & Test \\
\hline
{ EfficientNetB0 } & { 32 } & { FALSE } & { 0.2 } & top-1 & 49.27 & 40.95 & 41.63 \\
& & & & top-5 & 76.65 & 68.49 & 67.21 \\
\cline{4-8}
& & & { 0.5 } & top-1 & 41.92 & 39.45 & 39.42 \\
& & & & top-5 & 69.61 & 68.03 & 67.02 \\
\cline{4-8}
& & & { 0.8 } & top-1 & 26.73 & 35.81 & 35.48 \\
& & & & top-5 & 54.29 & 63.67 & 64.71 \\
\cline{3-8}
& & { TRUE } & { 0.2 } & top-1 & 94.47 & 79.36 & 80.1 \\
& & & & top-5 & 99.73 & 93.36 & 94.33 \\
\cline{4-8}
& & & { 0.5 } & top-1 & 91.75 & 80.86 & 81.54 \\
& & & & top-5 & 99.34 & 94.27 & 95.1 \\
\cline{4-8}
& & & { 0.8 } & top-1 & 84.27 & 79.69 & 81.06 \\
& & & & top-5 & 97.13 & 94.47 & 95.19 \\
\cline{2-8}
& { 16 } & { FALSE } & { 0.2 } & top-1 & 51.12 & 42.53 & 42.5 \\
& & & & top-5 & 78.90 & 69.07 & 69.62 \\
\cline{4-8}
& & & { 0.5 } & top-1 & 40.89 & 40.21 & 38.56 \\
& & & & top-5 & 69.69 & 67.85 & 66.92 \\
\cline{4-8}
& & & { 0.8 } & top-1 & 26.35 & 36.92 & 36.54 \\
& & & & top-5 & 52.73 & 65.27 & 63.75 \\
\cline{3-8}
& & { TRUE } & { 0.2 } & top-1 & 95.01 & 81.31 & \textbf{83.46} \\
& & & & top-5 & 99.83 & 94.78 & \textbf{94.9} \\
\cline{4-8}
& & & { 0.5 } & top-1 & 87.78 & 80.61 & 81.35 \\
& & & & top-5 & 98.76 & 94.46 & 94.52 \\
\cline{4-8}
& & & { 0.8 } & top-1 & 76.48 & 74.94 & 74.62 \\
& & & & top-5 & 94.24 & 91.56 & 92.02 \\
\hline
\end{tabular}}
\label{tab4}
\end{table}

Same as other models, the EfficientNet-B0 result is better when fine-tuned. Furthermore, the test accuracy result from EfficientNet-B0 from a batch size of 16 and 32 does not result in a significant difference.

\begin{table}[!h]
\caption{EfficientNetB7 Classification Results}
\label{table}
\setlength{\tabcolsep}{3pt}
\renewcommand{\arraystretch}{1.1}
\resizebox{.48\textwidth}{!}{
\begin{tabular}{|c|c|c|c|c|c|c|c|}
\hline
Model & Batch Size & Fine Tune & Dropout Rate & Accuracy(\%) & Train & Valid & Test \\
\hline
{ EfficientNetB7 } & { 32 } & { FALSE } & { 0.2 } & top-1 & 77.31 & 50.78 & 52.17 \\
& & & & top-5 & 93.54 & 75.72 & 76.13 \\
\cline{4-8}
& & & { 0.5 } & top-1 & 66.78 & 49.87 & 51.01 \\
& & & & top-5 & 88.83 & 74.67 & 74.88 \\
\cline{4-8}
& & & { 0.8 } & top-1 & 24.08 & 30.27 & 30.7 \\
& & & & top-5 & 48.63 & 57.1 & 56.59 \\
\cline{2-8}
& { 16 } & { FALSE } & { 0.2 } & top-1 & 42.66 & 34.28 & 34.84 \\
& & & & top-5 & 69.37 & 62.11 & 62.27 \\
\cline{4-8}
& & & { 0.5 } & top-1 & 35.87 & 33.25 & 35.71 \\
& & & & top-5 & 62.96 & 61.15 & 60.35 \\
\cline{4-8}
& & & { 0.8 } & top-1 & 23.76 & 30.35 & 31.18 \\
& & & & top-5 & 49.44 & 57.6 & 57.75 \\
\cline{3-8}
& & { TRUE } & { 0.2 } & top-1 & 98.31 & 84.21 & 84.7 \\
& & & & top-5 & 99.97 & 95.23 & 95.96 \\
\cline{4-8}
& & & { 0.5 } & top-1 & 96.4 & 84.79 & 84.6 \\
& & & & top-5 & 99.94 & 95.43 & 96.15 \\
\cline{4-8}
& & & { 0.8 } & top-1 & 94.77 & 85.18 & \textbf{85.08} \\
& & & & top-5 & 99.73 & 95.88 & \textbf{95.96} \\
\hline
\end{tabular}}
\label{tab5}
\end{table}

Unfortunately, only EfficientNet-B7 does not have a batch size of 32 with fine-tune result because the computer that was used to test was not able to handle the considerable memory requirement. When fine-tuned was off, an overall batch size of 32 gave a better result in test accuracy. This might be because a bigger batch size results in less noise. In a batch size of 16, the results of the fine-tuned model were better than the non-fine-tuned one because a few top layers from the frozen base model were unfrozen. It is also observed that drop-out rates of 0.2, 0.5, and 0.8 are all suitable for fine-tuned models on a batch size of 16 since there is no significant difference in the results on the test accuracy. Unfortunately, this model is prone to overfit when fine-tuned because the training accuracy is near 100\%. 

\begin{table}[!h]
\caption{Test Results Comparison}
\label{table}
\setlength{\arrayrulewidth}{0.1mm}
\setlength{\tabcolsep}{2.5pt}
\renewcommand{\arraystretch}{3}
\resizebox{.48\textwidth}{!}{
\begin{tabular}{|c|c|c|c|c|}
\hline
Model & Number of Parameter & top-1 Accuracy (\%) & top-5 Accuray (\%) & Inference Time (ms) \\
\hline
\setlength{\tabcolsep}{1pt}
Inception-V3 & 22,069,154 & 76.73 & 90.87 & 29.83 \\
\hline
InceptionResnet-V2 & 54,536,546 & 74.98 & 90.76 & 70.26 \\
\hline
Mobilenet-V2 & 2,424,514 & 81.92 & 94.62 & 17.32 \\
\hline
EfficientNet-B0 & 4,216,094 & 83.46 & 94.9 & 33.68 \\
\hline
EfficientNet-B7 & 64,430,610 & 85.08 & 95.96 & 104.91 \\
\hline
\end{tabular}}
\label{tab6}
\end{table}

Test results show that in accuracy, EfficientNet-B7 outperforms other models that were tested. Slightly less accurate than EfficientNet-B7 is Mobilenet-V2 with 81.92\% in top-1 accuracy.

\begin{figure}[!h]
  \includegraphics[width=\linewidth]{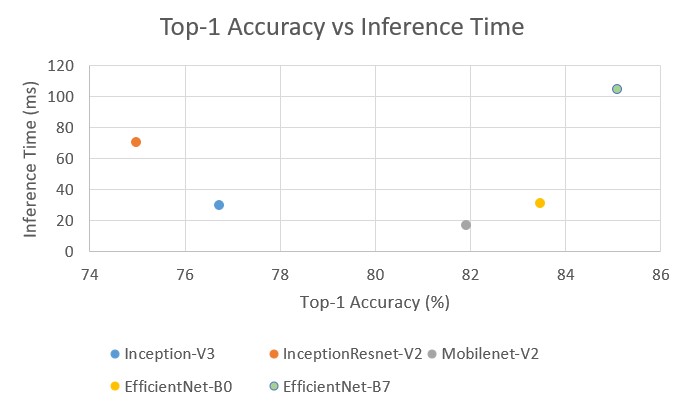}
  \caption{Top-1 Accuracy vs Inference Time.}
  \label{fig26}
\end{figure}

Mobilenet-V2 uses much fewer parameters and has the fastest inference time than EfficientNet-B7, with only 2.424.514 parameters and 17.32 ms inference time; the only catch is that Mobilenet-V2 is overfitting from the result in table 4. EfficientNet-B0 is more accurate and slightly slower than MobileNet-V2 but does not overfit as much as MobileNet-V2, as seen in table 5. Mobilenet-V2 and base EfficientNet-B0 are based on depthwise separable convolutions, requiring fewer parameters. In the case of Inception-V3, it can surpass Inception Resnet-V2 by a bit while having half the number of parameters. 

\subsection{Few-Shot Prototypical Network}
The following table below shows accuracy results from four different combinations of numbers of classes and images with prototypical networks.

\begin{table}[!h]
\caption{Prototypical Network Result}
\label{table}
\setlength{\arrayrulewidth}{0.1mm}
\setlength{\tabcolsep}{18pt}
\renewcommand{\arraystretch}{1.5}
\resizebox{.48\textwidth}{!}{
\begin{tabular}{|c|c|c|}
\hline
Accuracy (\%) & 5-way & 20-way \\
\hline
1-shot & 75.52 & 65.40 \\
\hline
5-shot & 89.27 & 80.52 \\
\hline
\end{tabular}}
\label{tab7}
\end{table}

The results show that accuracy with a few-shot scenario gave pretty good results, especially in a 5-way 5-shot scenario with 89.27\% accuracy. This shows that in a Japanese animation-style cartoon face dataset with only a given small number of examples, in this case, five examples in each class, it could accurately classify images in the query set. However, as the number of shots decreased and the number of ways or classes increased, the accuracy started to deteriorate. This is because prototypical networks only use mean vectors, so when the support set’s class variation is high and the number of examples is small, it might lead to misclassification.

\section{Conclusion}
\label {sec:conclusion}
This paper shows the use of various models to classify Japanese-styled character images. Mainly transfer learning and prototypical networks in which the dataset used differs in splitting the training, validation, and testing data. Prototypical networks use different classes for training, validation, and testing instead of all available classes.

The result shows that one of the transfer learning models, Efficient Net-B7 outperformed all other transfer learning models, with the top-1 and top-5 accuracy of 85.08\% and  95.96\%. However, this comes at the cost of the excessive number of parameters required and high inference time.

Another transfer learning model compensates for these drawbacks, Mobilenet-V2, which has a lower accuracy of top-1 81.92\% and top-5 94.62\% but required much fewer parameters than all of the other transfer learning models. This also results in a shallow inference time of only 17.32ms. However, this model also has its drawbacks, mainly the high risk of overfitting. On the other hand, EfficientNet-B0 fixed the overfitting problem in MobileNet-V2 but came with a cost of a bit more inference time and a few more parameters. 

A new way of tackling this image classification problem is using a few-shot method that uses far fewer images for the dataset, resulting in a more scalable model to add new classes. One of the models that can be used for few-shot learning is a prototypical network that uses a meta-learning algorithm to give an accurate estimation further.

Experimental results show that a 5-way 5-shot scenario prototypical networks produce a high accuracy of 89.27\%. This is because the number of classes is minimal, causing the model to easily differentiate between classes, even with a small image sample.

However, because of the use of mean vectors, the accuracy began to plummet significantly as the number of shots decreased and the number of ways or classes increased. 

The result shows that the prototypical network is a viable alternative to traditional transfer learning models regarding Japanese-styled character image classification. Mainly because of the small number of images needed for a quicker training process.

Future works are expected to include a future few-shot learning framework that might overcome the misclassification problem with prototypical networks.

\bibliographystyle{unsrt}
\bibliography{sample-bibliography}

\end{document}